\documentclass[journal]{IEEEtran}
\usepackage{amsmath}
\usepackage{amssymb}
\usepackage{mathtools,amsthm}
\usepackage{comment}
\usepackage{graphicx}
\usepackage{caption}
\usepackage{subcaption}
\usepackage{hyperref}
\usepackage{xcolor}
\usepackage{bbold}

\theoremstyle{definition}

\newtheorem{proposition}{Proposition}
\usepackage[ruled,linesnumbered]{algorithm2e}
\usepackage{enumitem}

\begin{document}

\title{Vision Language Model-Empowered Contract Theory for AIGC Task Allocation in Teleoperation}

\markboth{Journal of \LaTeX\ Class Files,~Vol.~14, No.~8, August~2015}%
{Shell \MakeLowercase{\textit{et al.}}: Bare Demo of IEEEtran.cls for IEEE Journals}

\author{Zijun~Zhan,
        Yaxian~Dong,
        Daniel~Mawunyo~Doe,
        Yuqing~Hu,
        Shuai~Li,
        Shaohua Cao,
        and~Zhu~Han,~\IEEEmembership{Fellow,~IEEE}
}

\maketitle

\begin{abstract}
Integrating low-light image enhancement techniques, in which diffusion-based AI-generated content (AIGC) models are promising, is necessary to enhance nighttime teleoperation. Remarkably, the AIGC model is computation-intensive, thus necessitating the allocation of AIGC tasks to edge servers with ample computational resources. Given the distinct cost of the AIGC model trained with varying-sized datasets and AIGC tasks possessing disparate demand, it is imperative to formulate a differential pricing strategy to optimize the utility of teleoperators and edge servers concurrently. Nonetheless, the pricing strategy formulation is under information asymmetry, i.e., the demand (e.g., the difficulty level of AIGC tasks and their distribution) of AIGC tasks is hidden information to edge servers. Additionally, manually assessing the difficulty level of AIGC tasks is tedious and unnecessary for teleoperators. To this end, we devise a framework of AIGC task allocation assisted by the Vision Language Model (VLM)-empowered contract theory, which includes two components: VLM-empowered difficulty assessment and contract theory-assisted AIGC task allocation. The first component enables automatic and accurate AIGC task difficulty assessment. The second component is capable of formulating the pricing strategy for edge servers under information asymmetry,  thereby optimizing the utility of both edge servers and teleoperators. The simulation results demonstrated that our proposed framework can improve the average utility of teleoperators and edge servers by $10.88 \sim 12.43\% $ and $1.4 \sim 2.17\% $, respectively. Code and data are available at https://github.com/ZiJun0819/VLM-Contract-Theory.
\end{abstract}


\IEEEpeerreviewmaketitle

\section{Introduction}

\begin{figure}[!t]
    \begin{center}
        \includegraphics[width=\linewidth]{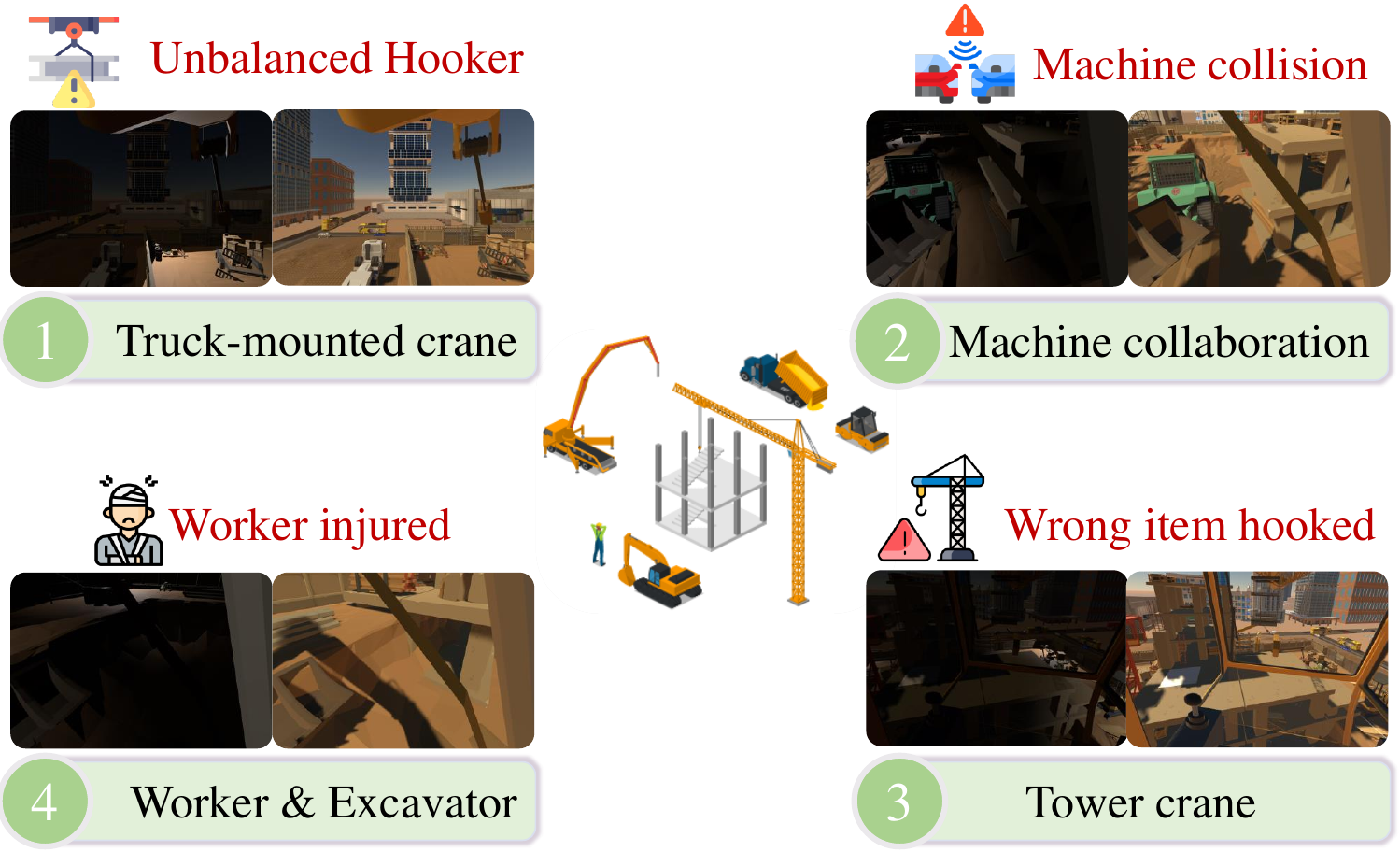}
        \caption{Potential consequences for teleoperators operating in low-light condition. In each item, the left sub-figure depicts a low-light image, while the right sub-figure shows the corresponding normal-light image.}
        \label{fig1}
    \end{center}
\end{figure}

\IEEEPARstart{D}{riven} by the demands of increasing productivity, reducing operational risks, and saving overhead \cite{zhan2023}, teleoperation has gained significant traction in recent years. However, the practical application of teleoperation still faces certain challenges, particularly in low-light conditions like nighttime construction sites, tunnels, and sewers \cite{wang2023improved}. Concretely, teleoperation needs to accomplish real-time interaction between the video streams from the construction site and the operation commands issued by the teleoperator. At night, the images captured by the camera will suffer from uneven illumination, low contrast, and color distortion \cite{chen2024}, rendering it challenging for the teleoperator to issue precise operation commands, thereby potentially leading to several serious consequences as illustrated in Fig. \ref{fig1}.

To refine teleoperation, it is essential to integrate low-light image enhancement techniques. Mainstream low-light image enhancement techniques include supervised deep learning models and unsupervised deep learning models \cite{fu2022}. In which, unsupervised low-light image enhancement models can subtly tackle the issue of collecting paired datasets; however, certain semantic information will lost when performing image enhancement, thereby compromising the downstream visual task \cite{zheng2023learning}. In contrast, benefiting from labeling information, supervised low-light image enhancement models can consistently maintain a more stable and excellent performance \cite{jiang2023low}, such as the state-of-the-art diffusion model \cite{dhariwal2021diffusion}.

Notably, the diffusion model is computationally intensive and the tele-robots are typically not equipped with ample computational resources, thus posing a challenge to accomplish the low light image enhancement task within the response time requirement. To this end, inspired by the AI-generated content (AIGC) edge architecture proposed by \cite{du2024enabling}, we propose offloading the low-light image enhancement task (AIGC task) to edge servers equipped with the diffusion-based AIGC model, so as to ensure real-time and high-quality interactive teleoperation.

\begin{figure}[!t]
    \begin{center}
        \includegraphics[width=\linewidth]{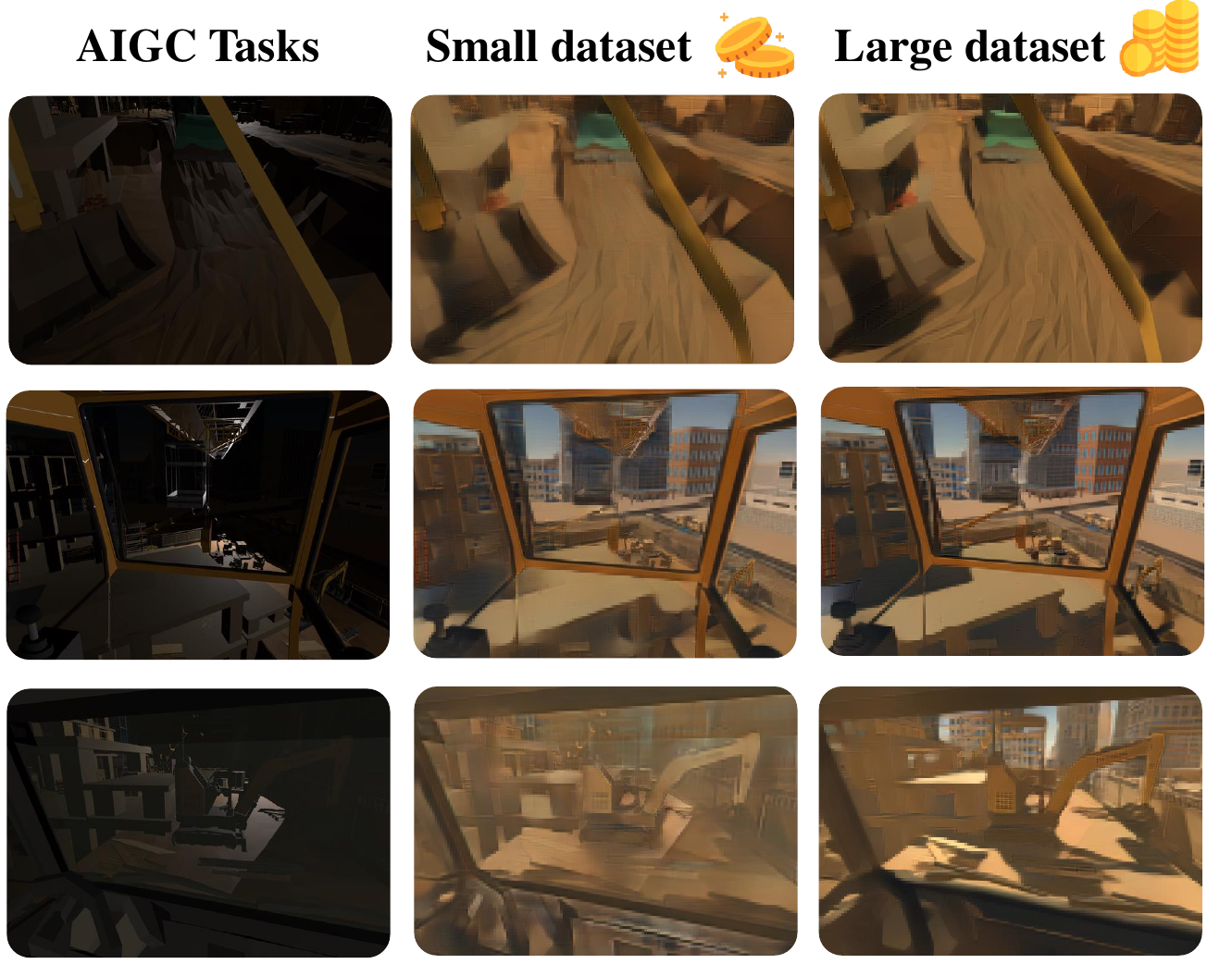}
        \caption{Results of AIGC tasks by diffusion-based AIGC models that are trained with varying-sized datasets.}
        \label{fig2}
    \end{center}
\end{figure}

Regarding AIGC task offloading, teleoperators should charge for the AIGC service to incentivize the edge server to perform the AIGC task \cite{wang2023incentive}. Remarkably, to guarantee the efficient offloading of AIGC tasks, varying difficulty of AIGC tasks should be handled by different types of AIGC models. As depicted in Fig. \ref{fig2}, in terms of visual perception, the first and second AIGC tasks can be processed effectively via an AIGC model trained on a small dataset, while the third one requires an AIGC model trained on a large dataset. Accordingly, given paired training data collection and model training, the cost of the AIGC model trained with varying-sized datasets is distinct. Therefore, it is crucial for edge servers to formulate differentiated pricing strategies to safeguard the utility of both edge servers and teleoperators. However, edge servers will confront information asymmetry in formulating pricing strategies, i.e., the difficulty level of AIGC tasks and their distribution are hidden information to edge servers before receiving offloaded AIGC tasks, thereby posing a challenge to identifying the optimal pricing strategy. For this reason, the first research question of this paper is 
\begin{itemize}
    \item[Q1:] Regarding edge servers, how can a differential pricing strategy for AIGC models be formulated under information asymmetry, so as to optimize the utility of both edge servers and teleoperators?
\end{itemize}

For Q1, certain solutions exist that utilize contract theory to solve similar problems \cite{huang2021efficient}\cite{li2022joint}. For instance, since the offloading service providers do not know the actual computing resources of the parked vehicles, i.e., information asymmetry, the authors in \cite{huang2021efficient} utilized contract theory to formulate differential pricing strategies to optimize the utility of the offloading system.  In addition, the authors in \cite{li2022joint} proposed a contract theory-empowered resource pricing scheme in the context of information asymmetry where the offloading service requesting vehicle is not aware of the free resources and dwell time of the offloading service providing vehicles. The contract bundles of previously mentioned contract theory-empowered pricing schemes are generally in the form of \textbf{\{}computational resources, pricing\textbf{\}}, all of which are quantifiable metrics. However, when using contract theory to solve Q1 results in a contract bundle of \textbf{\{}AIGC model type, pricing\textbf{\}}, in which, different from the computational resources, the AIGC model in the contract bundle is challenging to quantify. Therefore, we modify the contract bundle associated with Q1 to \textbf{\{}required performance of the AIGC model, pricing\textbf{\}}. Here, the contract bundle is mapped with the AIGC task difficulty level, thus guaranteeing the utility of both edge servers and teleoperators.

\begin{figure}[!t]
    \begin{center}
        \includegraphics[width=\linewidth]{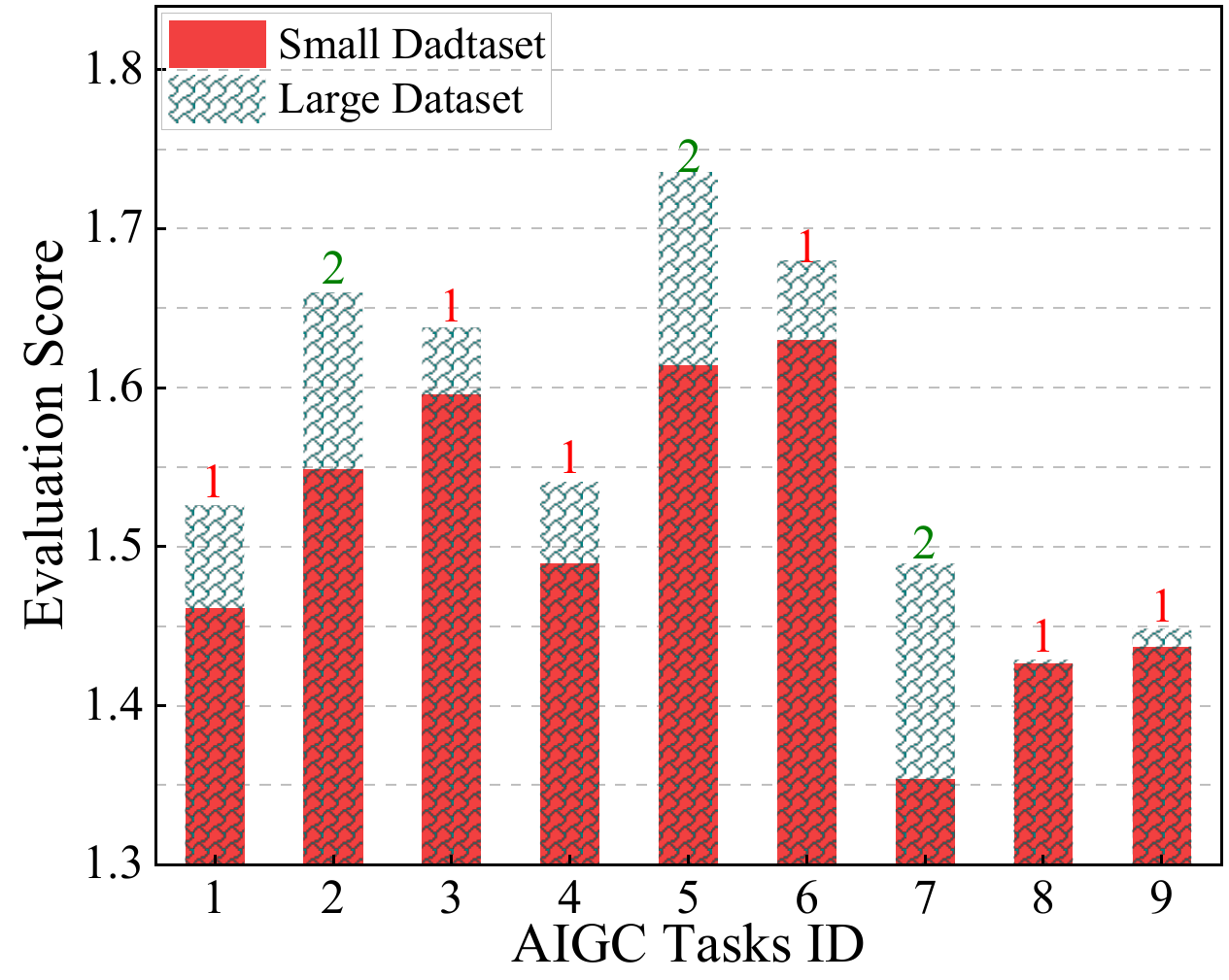}
        \caption{Evaluation score of AIGC tasks on AIGC models trained with varying-sized datasets. The evaluation score is the addition of the LPIPS \cite{zhang2018unreasonable} and SSIM \cite{wang2004image} value of processed results.}
        \label{fig3}
    \end{center}
\end{figure}

Upon pricing strategy formulated by edge servers, the teleoperator should opt for AIGC models according to the difficulty level of AIGC tasks. As shown in Fig. \ref{fig3}, considering the performance scores of different AIGC tasks processed by models trained on various-sized datasets, not all AIGC tasks should leverage the AIGC model trained on large datasets. The optimal strategy is for the teleoperator to determine the AIGC service based on the difficulty of the AIGC task. Nonetheless, manually identifying the difficulty level of AIGC tasks might be tedious and unnecessary for workers, as the envision of the teleoperation system is to enable workers to operate with a daytime-like experience even at night, without additional effort. In other words, the vision information received by workers at night should closely resemble that obtained under normal lighting conditions, thus leading to the second research question
\begin{itemize}
    \item[Q2:] Regarding teleoperators, how can the difficulty level of AIGC tasks be assessed automatically and accurately to ensure teleoperator efficiency?
\end{itemize}

Inspired by the use of LLaVA, a trained Visual Large Language Model based on LLaMA, to simulate five categories of realistic personalities for subjectively rating different AI-generated images \cite{du2023user}, we intend to utilize the Vision Language Model (VLM) to aid teleoperators in automatically assessing the difficulty level of AIGC tasks, thereby addressing Q2. The VLM-aided difficulty evaluation consists of three processes: evaluation metrics determination, initialization guidance, and prompt determination. First, the prompt engineer will interact with the VLM agents to identify difficulty evaluation metrics and conduct statistics on the accuracy of the difficulty evaluation of AIGC tasks under different metrics, thereby determining the optimal evaluation metrics. Next, upon the determined evaluation metrics, the VLM agents will be initialized and instructed. Finally, the prompt engineer will determine the final prompt based on the evaluation metrics and use it to interact with the VLM agent to assess the difficulty of various AIGC tasks.

\begin{figure*}[!t]
    \begin{center}
        \includegraphics[width=\linewidth]{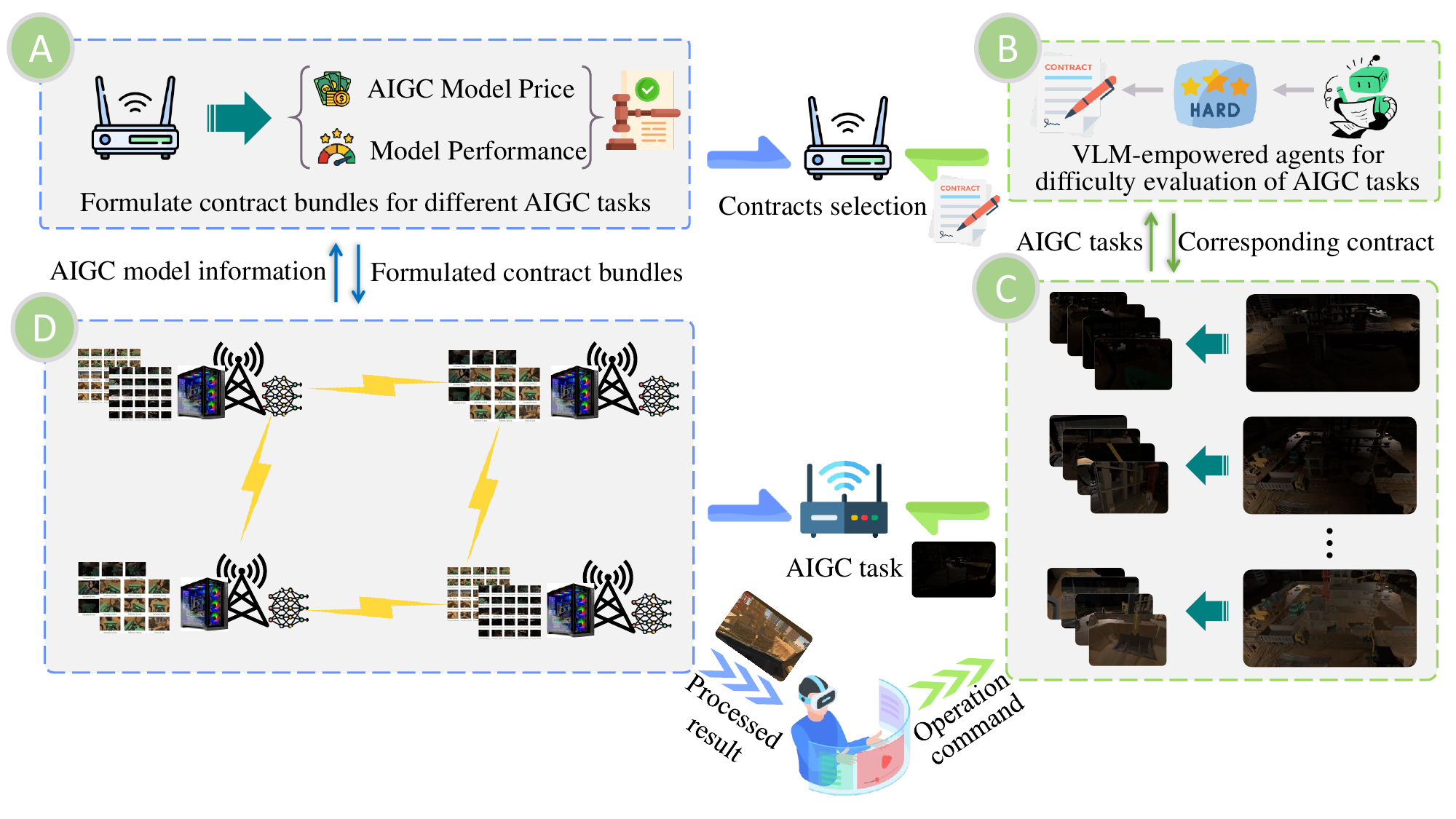}
        \caption{Framework overview of AIGC task allocation assisted by VLM-empowered contract theory in teleoperation.}
        \label{fig4}
    \end{center}
\end{figure*}

By integrating the solution of Q1 and Q2, our proposed framework of VLM-empowered contract theory for assisting AIGC task offloading is presented in Fig. \ref{fig4}. The solutions of Q1 and Q2 are illustrated in parts A and B, respectively. In Part C, various AIGC tasks will be generated for different nighttime construction sites and uploaded to the smart gateway for task offloading preparation via wireless transmission. In Part D, AIGC models trained on datasets of varying sizes will be deployed on edge servers, which are responsible for handling different AIGC tasks. Notably, when Part D is completed, the edge server will return the processed results directly to the teleoperator, who will then issue precise operation commands based on the received results. The main contributions of this paper are summarized as follows

\begin{enumerate}
    \item[1.] A framework of VLM-empowered contract theory for AIGC task allocation in the teleoperation scenario and the mathematical model of this framework are proposed. Under information asymmetry, the edge server has incomplete information on the difficulty level of AIGC tasks, this framework can guarantee the utility of workers and edge servers simultaneously.
    \item[2.] We present an automatic and accurate AIGC task difficulty level evaluation scheme by using VLM-empowered generative agents, thus ensuring the automation of contract theory-enabled task allocation. With meticulously crafted prompts, generative agents can effectively assess the difficulty level of various AIGC tasks, thereby ensuring the utility of teleoperators.
    \item[3.] Numerical simulation experiments have been done. The results showcased that our proposed framework is capable of augmenting the average utility of teleoperators and edge servers $10.88 \sim 12.43 \%$ and $1.4 \sim 2.17 \%$, respectively.
\end{enumerate}

The rest of this paper is structured as follows: Section \ref{sec:2} reviews related literature. In Sections \ref{sec:3} and \ref{sec:4}, we pinpoint the framework and system model of AIGC task allocation assisted by VLM-empowered contract theory. Section \ref{sec:5} details the VLM-empowered contract theory and the contract theory-assisted solutions for AIGC task allocation. Section \ref{sec:6} evaluates the proposed framework and VLM-empowered contract theory. The paper concludes with Section \ref{sec:7}.

\section{Related Works} \label{sec:2}
In this Section, we review several related works, which are AIGC service in networks and contract theory for task allocation, respectively.

\subsection{AIGC Service in Networks} \label{sec:2.1}
With the proliferation of Large Language Models, numerous studies have explored the integration of AIGC services with edge networks. The authors in \cite{wang2024guiding} introduced a framework that integrates wireless perception with AIGC, enhancing the quality of generated digital content and better aligning it with user preferences. In \cite{du2023user}, the authors proposed an LLM-empowered reinforcement learning method aimed at maximizing the subjective Quality of Experience and energy efficiency for users adopting Generative Diffusion Model (GDM)-based AIGC services for image generation. In \cite{fan2023learning}, AIGC services are considered computation-intensive, and the authors devised a decentralized incentive mechanism for AIGC service allocation. Specifically, they utilized a multi-agent deep reinforcement learning method to balance the supply and demand between AIGC services and vehicles, optimizing user experience and reducing transmission latency. Given the stochastic nature of AIGC services, the authors in \cite{du2024enabling} formulated an AIGC service provider (ASP) selection problem to enhance user satisfaction and employed deep reinforcement learning to match users with the optimal ASP. Analogously, in \cite{du2024diffusion}, the authors presented an AI-generated optimal decision algorithm for ASP selection. To facilitate the operation of AIGC services on resource-constrained mobile devices, the authors in \cite{du2023exploring} proposed a collaborative distributed diffusion-based AIGC framework. This framework distributes the computational workload of the diffusion model's reverse process among multiple mobile devices, thereby effectively reducing the overhead of running the AIGC service.

In addition to the above works, several studies focus on the data generated via AIGC service in edge networks. For instance, the authors in \cite{liu2024blockchain} proposed a blockchain-empowered AIGC framework and a consensus protocol, named proof of AIGC, to protect the copyright and ownership of AIGC. The authors in \cite{lin2023blockchain} addressed the trust issues of AIGC in the metaverse by designing a smart contract-based verification mechanism to ensure the authenticity of AIGC. Similarly, the authors in \cite{wang2023security} also focused on the trustworthiness of AIGC, presenting a series of countermeasures to enhance the security and privacy of AIGC-generated data.

These works, from various perspectives, investigate AIGC services in networks and pave the way for future research in this domain. Notably, distinct from existing studies, our work focuses on addressing AIGC service allocation from an information asymmetry perspective.

\subsection{Contract Theory for Task Allocation} \label{sec:2.2}
Contract theory excels in tackling the information asymmetry problem by aligning the incentive of agents with the utility of principals, which has been extensively investigated in the wireless communication sector for task allocation \cite{zhang2017contract}. For example, in the context of Multi-access Edge Computing (MEC) having limited knowledge of how much effort the unmanned area vehicles (UAVs) will exert for resource sharing, the authors in \cite{dang2023contract} proposed a contract theory-based mechanism to spur the UAVs.  The authors in \cite{zheng2023contract} leveraged contract theory to resolve the information asymmetry encountered during resource allocation between the source transmitter and relay nodes. In which, the information asymmetry indicates the source transmitter is unknown the computation time and relay power of relay nodes. In the context of task offloading, the delay tolerance and sensitivity of users are unknown for edge servers, thus, the authors in \cite{diamanti2022incentive} utilized contract theory to augment the resource utilization efficiency.

In addition to the aforementioned wireless scenarios, the application of contract theory for task allocation in vehicular edge computing has also been extensively investigated. In distributed vehicular edge computing, where the computing capability and energy resources of surrounding vehicles (SuVs) are unknown to task vehicles (TaVs), the authors in \cite{shen2023dynamic} devised an exploration and exploitation-assisted contract theory method to motivate SuVs to perform offloaded tasks from TaVs. In the framework of Collaborative Vehicular Edge Computing, the authors in \cite{huang2021efficient} used contract theory to encourage parked vehicles to share resources with other vehicles, despite the parking behaviors and computing overheads being unknown to other vehicles, causing information asymmetry. Furthermore, since the roadside unit (RSU) has limited knowledge of vehicle parking durations, the authors in \cite{li2022joint}, \cite{yang2020resource}, \cite{wen2023task}, and \cite{kazmi2021novel} leveraged contract theory to incentivize vehicles to rent resources to the RSU, thereby optimizing various metrics during task allocation.

Inspired by these existing works, we leverage the contract theory to acquire the pricing strategy for edge servers, resolving the information asymmetry caused by edge servers unknown the difficulty level distribution of AIGC tasks. Notably, different from existing works, the contract bundle, i.e., the AIGC model, in our paper, is challenging to quantify. Furthermore, the contract theory utilized in this paper is aided by VLM, which is the first attempt as far as we know.

\section{Framework Illustration} \label{sec:3}
In this section, we introduce the framework of AIGC task allocation assisted by VLM-empowered contract theory.

\subsection{Framework Overview} \label{sec:3.1}
The AIGC task allocation framework proposed in this paper is depicted in Fig. \ref{fig4}, which contains four main components, namely, the contract theory in Part A, the VLM agents in Part B, the construction site in Part C, and the edge servers in Part D. Subsequently, we will elaborate the specific functions of each component.

\textbf{Contract Theory:} The contract bundles in the task allocation system will generated via the smart gateway by invoking the contract theory in real time. Concretely, the generation of the contract bundles relies collectively on the model information of all edge servers and the distribution of AIGC task difficulty level in the offloading system.

\textbf{VLM Agents:} Upon generation of an AIGC task, similar to \cite{du2023user}, the VLM agents will assess its difficulty based on the pre-defined prompt and select the corresponding contract bundle for it. Subsequently, the smart gateway will determine the optimal task allocation scheme based on the environmental information of the offloading system and the contract bundles selected by AIGC tasks.

\textbf{Construction Site:} Numerous AIGC tasks will generated at the construction site. Here, AIGC tasks are low-light images captured by cameras equipped with various types of machines on the construction site, such as excavators, cranes, and bulldozers. It is worth noting that, we assume that the difficulty level of all AIGC tasks is determined before offloading \footnote{The difficulty evaluation by VLM agents can be done in several seconds. During this time, the light condition will not vary significantly.}.

\textbf{Edge Servers:} Different edge servers will be equipped with diffusion-based AIGC models trained on varying-sized datasets. The AIGC tasks will be scheduled as per the task allocation algorithm deployed in the smart gateway. In addition, similar to \cite{cao2023delay}, we assume that the computational resources of the edge servers are indivisible, i.e., multiple AIGC tasks offloaded to the edge servers concurrently will be executed sequentially and incur a queuing delay.

\subsection{Framework Workflow} \label{sec:3.2}
\begin{figure}[!t]
    \begin{center}
        \includegraphics[width=\linewidth]{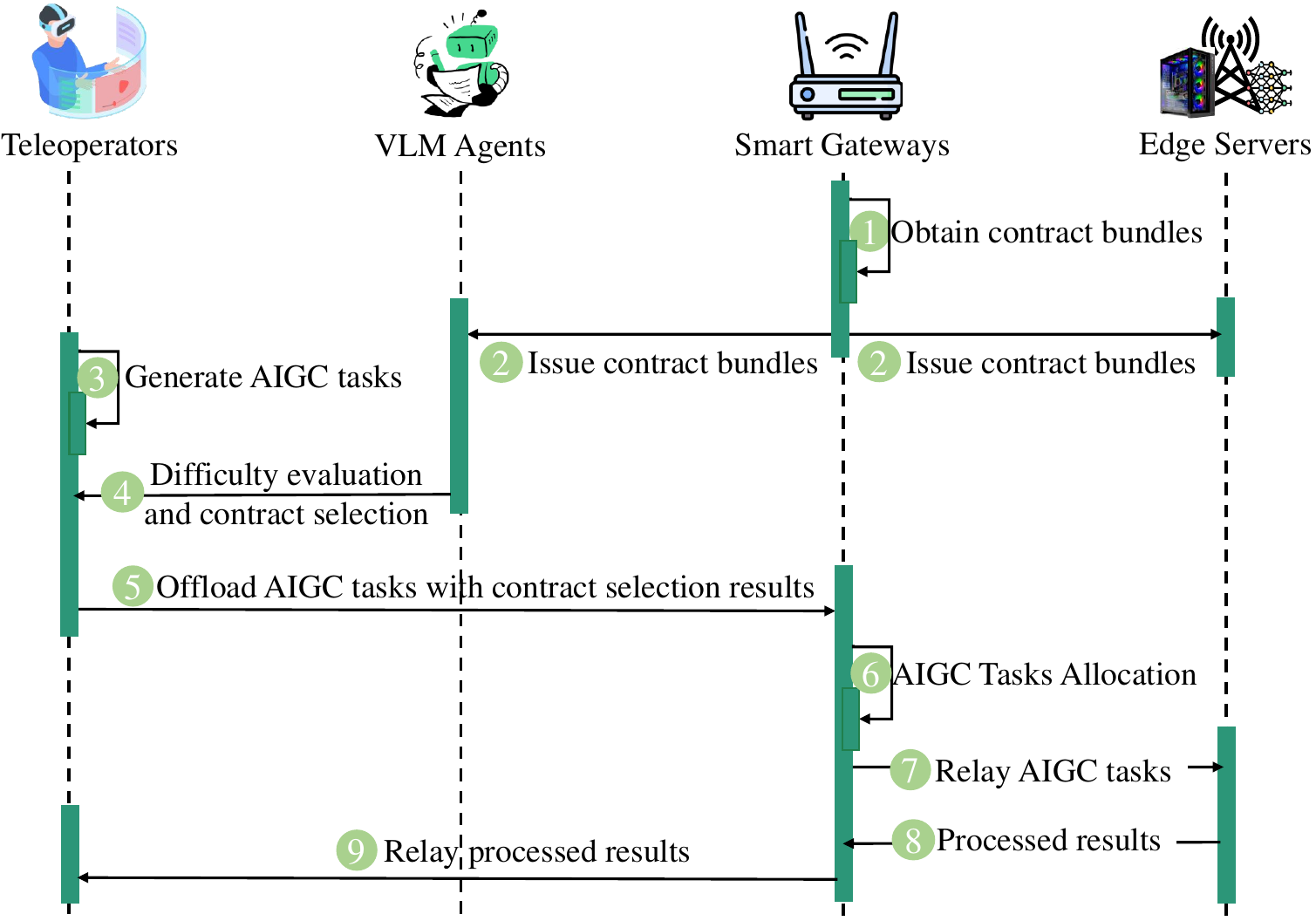}
        \caption{Flowchart of AIGC task allocation assisted by VLM-empowered contract theory in teleoperation.}
        \label{fig5}
    \end{center}
\end{figure}

To enhance the clarity of our proposed framework, we place the specific flowchart of the proposed framework in Fig. \ref{fig5}, encapsulating the interactions among four entities, namely, the teleoperator, the VLM agents, the smart gateway, and the edge server. Notably, the teleoperator will control machines on the construction site to generate various AIGC tasks.

\section{System Model} \label{sec:4}
As per the framework presented in Section \ref{sec:3}, we pinpoint the system model in three subsections, i.e., the model of AIGC service in Section \ref{sec:4.1}, the model of contract theory in Section \ref{sec:4.2}, and the model of task allocation in Section \ref{sec:4.3}.

\subsection{Model of AIGC Service} \label{sec:4.1}
In this paper, the AIGC service is provided by the diffusion-based AIGC model. The diffusion model includes two phases, i.e., forward process and reverse process \cite{ho2020denoising} \cite{song2021denoising}. 

The forward process is a diffusion process, in which the noise will gradually added to the original data and eventually obtain the pure noise that abides by Gaussian distribution. Mathematically, the final obtained pure noise $\mathbf{x}_T \sim \mathcal{N}(0, I)$ is progressively transformed from the original data $\mathbf{x}_0$ via
\begin{equation} \label{eq:1}
    q (\mathbf{x}_{T}|\mathbf{x}_0) = \prod_{t=1}^T q(\mathbf{x}_{t}|\mathbf{x}_{t-1})
\end{equation}
and 
\begin{equation} \label{eq:2}
    q(\mathbf{x}_{t}|\mathbf{x}_{t-1})=\mathcal{N}(\mathbf{x}_t;\sqrt{1-\beta_t}\mathbf{x}_{t-1}, \beta_t\mathbf{I})
\end{equation}
in $T$ steps. Here, $\beta_t$ is the schedule parameter of the diffusion model.

The reverse process, a.k.a. denoising process, is the key component of the diffusion model, also of the AIGC service, which aims to deduce the target $\hat{\mathbf{x}}_0$ from the noise sample $\hat{\mathbf{x}}_T \sim \mathcal{N}(0, \mathbf{I})$. i.e.,
\begin{equation} \label{eq:3}
    p_{\phi}(\hat{\mathbf{x}}_0) = p(\hat{\mathbf{x}}_T) \prod_{t=1}^T p_{\phi}(\hat{\mathbf{x}}_{t-1}|\hat{\mathbf{x}}_t),
\end{equation}
where $p(\hat{\mathbf{x}}_T)=\mathcal{N}(0, \mathbf{I})$.

Notably, since the AIGC service utilized in this paper is for low-light image enhancement, the reverse process will be a conditional denoising process \cite{jiang2023low}. Therefore, the Eq.(\ref{eq:3}) should be transformed into
\begin{equation} \label{eq:4}
    p_{\phi}(\hat{\mathbf{x}}_0) = p(\hat{\mathbf{x}}_T) \prod_{t=1}^T p_{\phi}(\hat{\mathbf{x}}_{t-1}|\hat{\mathbf{x}}_t, \tilde{x}),
\end{equation}
 with a conditional input $\tilde{x}$, i.e., low lighting condition image in this paper. Here, $p_{\phi}(\hat{\mathbf{x}}_{t-1}|\hat{\mathbf{x}}_t, \tilde{x})$ is similar to (\ref{eq:2}),
 \begin{equation} \label{eq:5}
     p_{\phi}(\hat{\mathbf{x}}_{t-1}|\hat{\mathbf{x}}_t, \tilde{x})=\mathcal{N}(\tilde{\mathbf{x}}_{t-1};\mathbf{\mu_\phi}(\hat{\mathbf{x}}_{t}, t, \tilde{\mathbf{x}}), \sigma^2_t\mathbf{I}),
 \end{equation}
where $\mathbf{\mu_\phi}(\hat{\mathbf{x}}_{t}, t, \tilde{\mathbf{x}})$ and $\sigma^2_t$ are approximated value of the noise added in the t-th step of the forward process and variance parameter, respectively. Here, $\mathbf{\mu_\phi}(\hat{\mathbf{x}}_{t}, t, \tilde{\mathbf{x}})$ is defined as
\begin{equation} \label{eq:6}
    \mathbf{\mu_\phi}(\hat{\mathbf{x}}_{t}, t, \tilde{\mathbf{x}})=\frac{1}{\sqrt{\alpha_t}}(\hat{\mathbf{x}}_{t} - \frac{\beta_t}{\sqrt{1 - \bar{\alpha}_t}} \epsilon_\phi(\hat{\mathbf{x}}_{t}, t, \tilde{\mathbf{x}})),
\end{equation}
where $\alpha_t = 1 - \beta_t, \bar{\alpha}_t=\prod_{1}^{t} \alpha_t$.

In summary, the kernel of the AIGC model is to train a network to achieve the function of $\epsilon_\phi(\hat{\mathbf{x}}_{t}, t, \tilde{\mathbf{x}})$. Therefore, the input and output of the AIGC service will be $\tilde{\mathbf{x}}$ and $\hat{\mathbf{x}}_{0}$, representing the low light condition image (AIGC task) and enhanced image (processed result), respectively.

\subsection{Model of Contract Theory} \label{sec:4.2}
When adopting the contract theory to formulate the pricing strategy for AIGC tasks with varying difficulty levels, hidden information exists, i.e., the difficulty level and associated distribution of AIGC tasks are unknown for edge servers, before offloading. Therefore, we begin by classifying the AIGC tasks by their difficulty levels.

For clarity, we consider the scenario of AIGC tasks with two levels of difficulty only: low and high, represented by $\theta_L$ and $\theta_H$, respectively. The distribution of these difficulty levels is denoted by the probabilities of $\theta_L$ and $\theta_H$ tasks, represented as $\beta_L$ and $\beta_H$, where $\beta_L + \beta_H = 1$. During AIGC task allocation, edge servers will predict these two parameters to adjust the contract bundle in real time.

For each type of AIGC task, represented by $\theta_i$, the pricing strategy is in the form of contract bundle $(p_i, I(\phi_i))$. $p_i$ means the price of the AIGC model tailored for $\theta_i$ AIGC tasks and $I(\phi_i)$ is required performance of the AIGC model.

\subsubsection{Utility model for edge servers} The utility model of edge servers depends on the payment purchased by teleoperators and the cost of data collection and model training. Since the cost of data collection and model training is proportion to the performance of the AIGC model, we utilize $I(\phi_i)$ to approximate the model cost. Therefore, we define the utility of edge servers with $\theta_i$ AIGC tasks as:
\begin{equation} \label{eq:7}
    {U_{{E_i}}} = {p_i} - {\eta _1}I({\phi _i}) + \Delta C,~ i \in \{ L,H\},
\end{equation}
where $\eta_1$ is the coefficient utilized to fine-tune the utility of edge servers. Since the revenue for compensating the cost of the AIGC model training has not been counted, we introduce $\Delta C$.

\subsubsection{Utility model for teleoperators} The utility model of teleoperators depends on the difficulty level of AIGC tasks and the quality and price of AIGC services. To this end, we define the utility of teleoperators with $\theta_i$ AIGC tasks as:
\begin{equation} \label{eq:8}
    {U_{{W_i}}} = {\theta _i}\ln [{\eta _2}(I({\phi _i}) - {I_{r1}})] + {\eta _3}(I({\phi _i}) - {I_{r2}}) - {p_i},~ i \in \{ L,H\},
\end{equation}
where $I_{r1}$ and $I_{r2}$ are the threshold and expected value of the AIGC model performance defined by teleoperators, in which $I_{r1} < I_{r2}$. $\eta_2$ and $\eta_3$ are the coefficients utilized to fine-tune the utility model of teleoperators.

\subsubsection{Contract theory problem formulation} Under information asymmetry, we need to solve (\ref{Problem_1}) to obtain the pricing strategy, i.e., the contract bundle, to maximize the utility of edge servers.
\begin{subequations}
	\begin{align}
		\max_{(p_i, I(\phi_i))} \quad &{\beta _L}({p_L} - {\eta _1}I({\phi _L})) + {\beta _H}({p_H} - {\eta _1}I({\phi _H})) + \Delta C \label{Problem_1} \\
		  s.t. \quad &U_{W_{i}}(p_i, I(\phi_i)) \ge 0, \label{P1_constraint_1}\\
            &U_{W_{i}}(p_i, I(\phi_i)) \ge U_{W_{i}}(p_j, I(\phi_j)), \label{P1_constraint_2}\\
            &\forall j \ne i,\quad i,j \in \{ L,H\}. \nonumber
	\end{align}
\end{subequations}
Here, (\ref{P1_constraint_1}) and (\ref{P1_constraint_2}) denote the Incentive Rationality (IR) and Incentive Compatibility (IC) constraints, respectively.

\subsection{Model of Task Allocation} \label{sec:4.3}
\subsubsection{System model for task allocation}
Suppose there are $N$ mutually independent AIGC tasks, $M$ edge servers, and $K$ smart gateways, defined as $\mathcal{T}=\{T_1,T_2,\cdots,T_n,\cdots,T_N\}$, $\mathcal{F}=\{F_1,F_2,\cdots,F_m,\cdots,F_M\}$, and $\mathcal{S}=\{S_1,S_2,\cdots,S_k,\cdots,S_K\}$, respectively. We utilize the binary offloading matrix $\mathbf{O}_{N \times M}$ to represent the task allocation scheme. In which, $o_{nm}=1$ indicates the AIGC task $T_n$ will offloaded to the edge server $F_m$.

In addition, similar to \cite{cao2023delay}\cite{azizi2022deadline}, we assume smart gateways are wire connected with edge servers and AIGC tasks will wirelessly transmit to the nearby smart gateway. The topology of smart gateways and edge servers is defined by $\mathcal{L}=\{e_{jl}| j,l \in \{\mathcal{S, F}\}\}$. In which, $e_{jl} = (e_{jl}^{p}, e_{jl}^{b})$, where $e_{jl}^{p}$ and $e_{jl}^{b}$ are the propagation time and bandwidth of the communication link. Moreover, it is worth noting that before allocation, the VLM agents will opt for the target edge servers for AIGC tasks as per their difficulty levels. We utilize $d_n = {1, 2}$ to represent the difficulty level of the AIGC task $T_n$. The target edge servers of $T_n$  is defined as $\mathcal{F}_1$ if $d_n = 1$, otherwise $\mathcal{F}_2$. For clarity, we utilize $\mathcal{F}_{d_n}$ to represent the target edge servers of $T_n$.

\subsubsection{Task response time}
As illustrated in Section \ref{sec:3}, the AIGC tasks will be wirelessly transmitted to smart gateways for offloading and the computational resource of edge servers is indivisible. Due to the construction site being proximate to the smart gateway, we only consider the transmission time from smart gateways to edge servers \cite{cao2023delay}. Therefore, the transmission time of the AIGC task $T_n$ is defined as
\begin{equation} \label{eq:10}
    \begin{aligned}
    d_{nm}^{t}=&\sum_{\forall e_{jl} \in \mathcal{L}}\left(2 \times \left(e_{jl}^{p}+\frac{\tilde{\mathbf{x}}}{e_{jl}^{b}}\right) \right) \times x_{jl}^{nm} \times o_{n m}, \\
    &\forall n \in \mathcal{T},m \in \mathcal{F}_{d_n},
\end{aligned}
\end{equation}
where $x_{i j}^{nm}=1$ means $T_n$ is offloaded via communication link $e_{jl}$ to $F_m$, otherwise $x_{i j}^{nm}=0$. $\tilde{\mathbf{x}}$ is the input of AIGC service mentioned in Section \ref{sec:4.1}.

The computation time of $T_n$ is defined as
\begin{equation} \label{eq:11}
    d_{nm}^{c}=\sum_{\forall m \in \mathcal{F}_{d_n}}\left(T_{n}^{r}/F_{m}^{c}\right) \times o_{n m}, \forall n \in \mathcal{T}.
\end{equation}
In which, $T_{n}^{r}$ is the computational resource required by the AIGC task $T_n$ and $F_{m}^{c}$ is the computational resource that the edge server $F_m$ possesses. Notably, $T_{n}^{r}$ is identical for all AIGC tasks since $\tilde{\mathbf{x}}$ is the same for all AIGC tasks.

Given the computational resources of edge servers are indivisible, multiple AIGC tasks offloaded to $F_m$ will cause queuing delay. Here, we define the queue of $F_m$ as $\mathcal{Q}_m$, and the the queuing time of $T_n$ is specified as
\begin{equation} \label{eq:12}
    d_{nm}^{q}=\sum_{\forall m \in \mathcal{F}_{d_n}} \sum_{\forall n^{\prime} \in Q_{m}}\left(T_{n^\prime}^{r} / F_{m}^{c}\right) \times o_{n m}, \forall n \in \mathcal{T}.
\end{equation}

In summary, the response time of AIGC task $T_n$ is the sum of transmission time, computation time, and queuing time, which is defined as
\begin{equation} \label{eq:13}
    D_{n}^{sum}=d_{nm}^{t}+d_{nm}^{c}+d_{nm}^{q}, \forall n \in \mathcal{T}, m \in \mathcal{F}_{d_n}.
\end{equation}

To evaluate whether AIGC tasks are completed on time, we introduce an auxiliary variable $s_n$ for each AIGC task $T_n$, which is specified as
\begin{equation} \label{eq:14}
    \left\{\begin{array}{l}
s_{n}=1, \text {if} \hspace{0.1cm} {D}_{n}^{sum}<\mathbb{D}, \\
s_{n}=0, \text {otherwise},
\end{array} \forall n \in \mathcal{T} ,m \in \mathcal{F}_{d_n},\right.
\end{equation}
where $\mathbb{D}$ is the required response time of AIGC tasks. With $D_{nm}^{sum}$ and $s_n$, we can obtain the completion rate and average response time of AIGC tasks as $R_{\%}=\sum_{\forall n \in \mathcal{T}} s_{n} / N$ and $\bar{\mathbb{D}}=\sum_{\forall n \in \mathcal{T}} {D}_{n}^{sum} / N$, respectively.

\subsubsection{Task Allocation Problem Formulation}
For AIGC task allocation, we intend to maximize the task completion rate and minimize the average response time of AIGC tasks. Therefore, we formulate the optimization problem below
\begin{subequations}
	\begin{align}
		\max_{\mathbf{O}} \quad &\zeta_1(1-R_{\%}) + \zeta_2\bar{\mathbb{D}} \label{Problem_2} \\
		  s.t. \quad &\sum_{m \in \mathcal{F}} o_{n m}=1,\quad \forall n \in \mathcal{T}, \label{P2_constraint_1}\\
            &x_{j l}^{nm} \in\{0,1\},\quad \forall j,l \in \{\mathcal{S},\mathcal{F}\}, \forall n \in \mathcal{T}, \forall m \in \mathcal{F}_{d_n}.\label{P2_constraint_2}
	\end{align}
\end{subequations}
$\zeta_1$ and $\zeta_2$ are the mapping coefficients to ensure that the two optimization objectives are in the same order of magnitude. (\ref{P2_constraint_1}) means that each AIGC task can be only offloaded to one edge server. (\ref{P2_constraint_2}) is the constraint on the communication link between the smart gateway and target edge server for offloading $T_n$.

\section{VLM-empowered Contract Theory for AIGC Task Allocation} \label{sec:5}
We begin with the specific procedure of AIGC tasks assessment in Section \ref{sec:5.1}. Subsequently, we present the solutions for problems (\ref{Problem_1}) and (\ref{Problem_2}) in Section \ref{sec:5.2}.
\subsection{Difficulty Evaluation via VLM-Empowered Agent} \label{sec:5.1}
\begin{figure}[!t]
    \begin{center}
        \includegraphics[width=\linewidth]{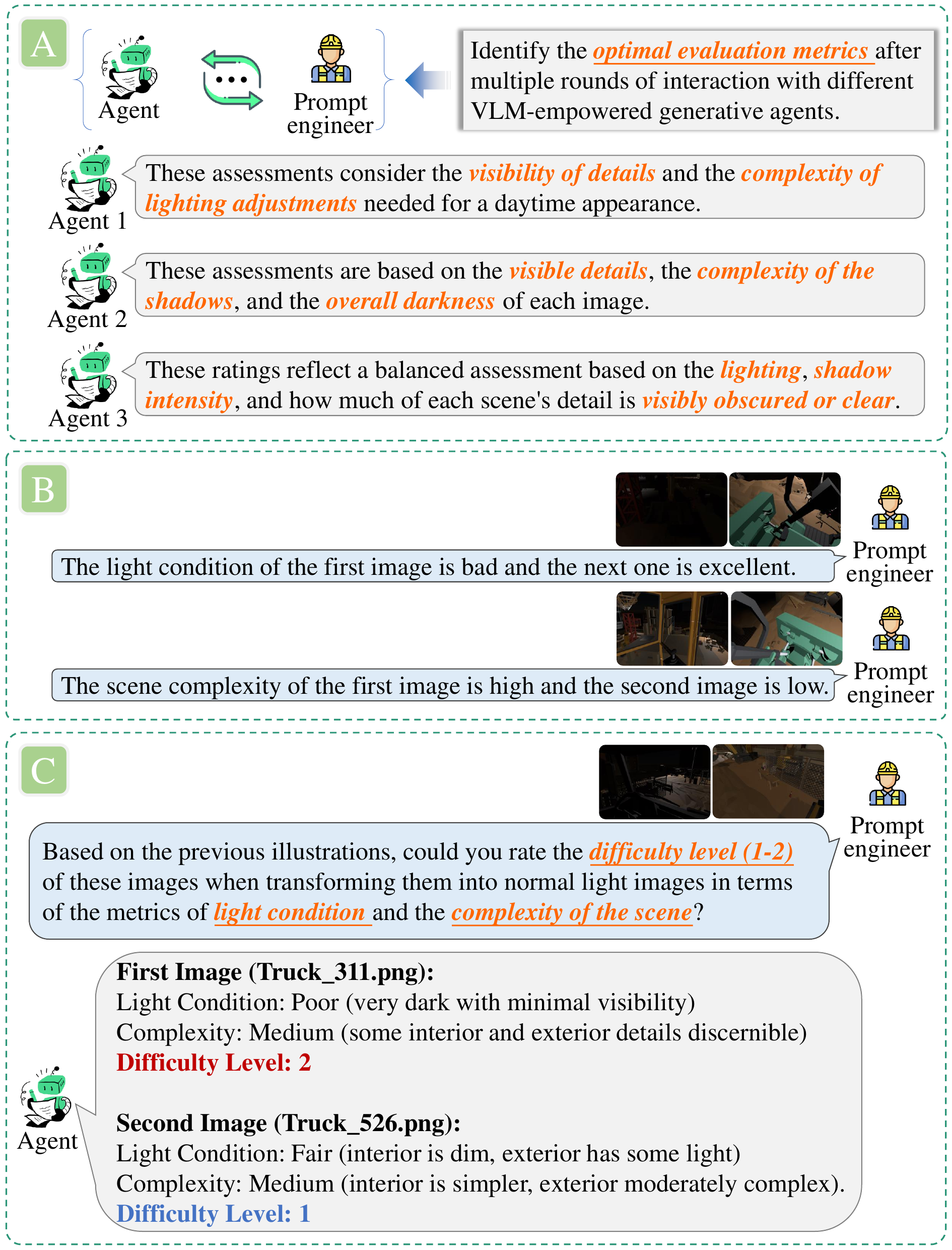}
        \caption{Procedure of AIGC tasks difficulty assessment via VLM-empowered generative agent.}
        \label{fig6}
    \end{center}
\end{figure}

The mechanism of our proposed VLM-empowered difficulty evaluation is analogous to the scheme mentioned in \cite{du2023user}, which utilizes the capability of chain-of-thought \cite{wei2022chain} of VLM agents. The specific procedure of our proposed method is depicted in Fig. \ref{fig6}, including three parts, optimal evaluation metrics determination in Part A, initial guidance for VLM agents in Part B, and difficulty evaluation in Part C.

As for Part A, the prompt engineer will begin by interacting with the VLM agent in multiple rounds only with the AIGC tasks, which difficulty level needs to be assessed. After each round of interaction, the prompt engineer elicited the reasoning information of VLM agents in the form of the output of agents 1, 2, and 3 depicted in Part A of Fig. \ref{fig6}. Subsequently, the prompt engineer will perform statistical analysis on the accuracy of VLM agents in AIGC tasks difficulty evaluation with different reasoning information and identify the optimal evaluation metrics.

With the optimal evaluation metrics obtained from Part A, the prompt engineer will provide several metrics-related examples to guide VLM agents. The basic rationale behind this step is similar to the essence of \cite{wei2022chain}, i.e., providing examples to refine the reasoning capability of VLM agents, so as to boost the performance of VLM agents.

Upon completion of Parts A and B of Fig. \ref{fig6}, the prompt engineer will develop the prompt according to determined optimal metrics. The prepared prompt and the specific difficulty assessment of the VLM agent to two distinct AIGC tasks are depicted in Part C of Fig. \ref{fig6}. Notably, the VLM agent not only assesses the AIGC tasks but also provides detailed reasoning information, aligning with the chain of thought prompting mentioned in \cite{wei2022chain}.

\subsection{Contract Theory-Assisted Task Allocation} \label{sec:5.2}
Regarding problem (\ref{Problem_1}), we can apply the optimization method mentioned in \cite{li2021contract} to reduce the constraint and derive the optimal contract bundle.

Concretely, after applying the optimization method, the four inequality constraints will be transformed into two equality constraints, which are written as
\begin{equation} \label{eq:16}
    {\theta _L}\ln [{\eta _2}(I({\phi _L}) - {I_{r1}})] + {\eta _3}(I({\phi _L}) - {I_{r2}}) - {p_L} = 0
\end{equation}
and
\begin{equation} \label{eq:17}
    \begin{array}{l}
        {\theta _H}\ln [{\eta _2}(I({\phi _H}) - {I_{r1}})] + {\eta _3}(I({\phi _H}) - {I_{r2}}) - {p_H}\\
         = {\theta _H}\ln [{\eta _2}(I({\phi _L}) - {I_{r1}})] + {\eta _3}(I({\phi _L}) - {I_{r2}}) - {p_L},
    \end{array}
\end{equation}
respectively.

After refining (\ref{eq:16}) and (\ref{eq:17}), we can acquire the expression of pricing for low and high difficulty levels AIGC tasks, i.e., $p_L$ and $p_H$, as
\begin{equation} \label{eq:18}
    {p_L} = {\theta _L}\ln [{\eta _2}(I({\phi _L}) - {I_{r1}})] + {\eta _3}(I({\phi _L}) - {I_{r2}})
\end{equation}
and
\begin{equation} \label{eq:19}
    {p_H} = {U_{W({\theta _H})}}(I({\phi _H})) - {U_{W({\theta _H})}}(I({\phi _L})) + {p_L}.
\end{equation}
In which, ${U_{W({\theta _H})}}(I({\phi _i}))$ is defined as
\begin{equation} \label{eq:20}
    {U_{W({\theta _H})}}(I({\phi _i})) = {\theta _H}\ln [{\eta _2}(I({\phi _i}) - {I_{r1}})] + {\eta _3}(I({\phi _i}) - {I_{r2}}).
\end{equation}

Subsequently, substituting (\ref{eq:18}) and \ref{eq:19} into problem (\ref{Problem_1}), the original constrained problem will be an unconstrained optimization problem, i.e.,
\begin{equation} \label{eq:21}
    {\beta _L}({p_L} - {\eta _1}I({\phi _L})) + {\beta _H}({p_H} - {\eta _1}I({\phi _H})) + \Delta C.
\end{equation}

To resolve the problem presented in (\ref{eq:21}) intuitively, we will derive its first and second derivations with respect to $I({\phi _LH})$ and $I({\phi _H})$ as
\begin{equation} \label{eq:22}
    \left\{ \begin{array}{l}
    \frac{{\partial U}}{{\partial I({\phi _L})}} = \frac{{{\theta _L} - {\beta _H}{\theta _H}}}{{I({\phi _L}) - {I_{r1}}}} + {\beta _L}({\eta _3} - {\eta _1}),\\
    \frac{{{\partial ^2}U}}{{\partial I{{({\phi _L})}^2}}} =  - \frac{{{\theta _L} - {\beta _H}{\theta _H}}}{{{{(I({\phi _L}) - {I_r})}^2}},}
    \end{array} \right.
\end{equation}
and
\begin{equation} \label{eq:23}
    \left\{ \begin{array}{l}
    \frac{{\partial {U_E}}}{{\partial I({\phi _H})}} = \frac{{{\beta _H}{\theta _H}}}{{I({\phi _H}) - {I_{r1}}}} + {\beta _H}({\eta _3} - {\eta _1}),\\
    \frac{{{\partial ^2}{U_E}}}{{\partial I{{({\phi _H})}^2}}} =  - \frac{{{\beta _H}{\theta _H}}}{{{{(I({\phi _H}) - {I_r})}^2}},}
    \end{array} \right.
\end{equation}
respectively.

Observing (\ref{eq:22}) and (\ref{eq:23}), we know the solution of problem (\ref{Problem_1}) exists with an assumption ${\theta _L} > {{\beta _H}{\theta _H}}$. With this assumption, the required performance of the AIGC model for low and high difficulty levels AIGC task, i.e., $I({\phi _L})$ and $I({\phi _h})$, are specified as
\begin{equation} \label{eq:24}
    I({\phi _L}) = {I_{r1}} + \frac{{{\theta _L} - {\beta _H}{\theta _H}}}{{{\beta _L}({\eta _1} - {\eta _3})}}
\end{equation}
and
\begin{equation} \label{eq:25}
    I({\phi _H}) = {I_{r1}} + \frac{{{\theta _H}}}{{{\eta _1} - {\eta _3}}},
\end{equation}
respectively.

Lastly, organizing (\ref{eq:18}), (\ref{eq:19}), (\ref{eq:24}), and (\ref{eq:25}), we can formulate the contract bundle, a.k.a the pricing strategy, for edge servers to provide differential AIGC service for teleoperator under information asymmetry.

It is worth noting that the formulated contract bundle possesses certain deficiencies. Specifically, the required performance of AIGC models, i.e., $I({\phi _L})$ and $I({\phi _H})$, cannot be promised, since $I({\phi _L})$ and $I({\phi _h})$ are fixed once the contract bundle is formulated and the actual performance of AIGC models are varying against AIGC tasks. Put another way, the AIGC task $T_n$ processed on the AIGC model trained with small and large datasets will purchase with $p_L$ and $p_H$ to acquire the actual performance score of $I_n({\phi _L})^{\prime}$ and $I_n({\phi _H})^{\prime}$, respectively. To this end, the difficulty assessment proposed in Section \ref{sec:5.1} should align as much as possible to the difficulty assessment solution obtained via revised contract bundle $(p_i, I_n(\phi _i)^{\prime})$, which we named as "Oracle Solution". The "Oracle Solution" is defined as
\begin{equation} \label{eq:26}
    \left\{ \begin{array}{l}
    {d_n} = 1, \text{if} ~ {U_{{W_H}}}({I_{{H^\prime }}},{p_H}) < {U_{{W_L}}}({I_{{L^\prime }}},{p_H}),\\
    {d_n} = 2, \text{otherwise},
    \end{array} \right.
\end{equation}
where $I_{{L^\prime }}$ and $I_{{H^\prime }}$ are interchangeable with $I_n({\phi _L})^{\prime}$ and $I_n({\phi _H})^{\prime}$ for the sake of clarity.

Organizing (\ref{eq:26}), we can obtain the following proposition,
\begin{proposition}\label{proposition_1}
\textit{The difficulty level of the AIGC tasks $T_n$ will assigned as 1, if the following inequality holds}
    \begin{equation} \label{eq:27}
		{\eta _2} < \frac{{{{\left( {\frac{{{\theta _H} - {\beta _H}{\theta _H}}}{{{\theta _L} - {\beta _H}{\theta _H}}} \cdot \frac{{{I_{{L^\prime }}} - {I_{{r_1}}}}}{{{I_{{H^\prime }}} - {I_{{r_1}}}}}} \right)}^{\frac{{{\theta _H}}}{{{\theta _H} - {\theta _L}}}}} \cdot {e^{\frac{{{\eta _3}}}{{{\beta _L}({\eta _1} - {\eta _3})}} - \frac{{{\eta _3}}}{{{\theta _H} - {\theta _L}}}({I_{{H^\prime }}} - {I_{{L^\prime }}})}}}}{{{I_{{L^\prime }}} - {I_{{r_1}}}}};
    \end{equation}
    \textit{Otherwise, the difficulty level should assigned as 2.}
\end{proposition}

Upon formulation of the contract bundle and AIGC tasks difficulty level assessment, we can tackle problem (\ref{Problem_2}) to derive the AIGC task allocation scheme. Notably, problem (\ref{Problem_2}) considers only the task allocation, which is similar to the problem formulated in \cite{cao2023delay}. Therefore, we can utilize the method proposed in their work to solve our problem as well, so we will not reiterate the solution here.

\textbf{Remark:} Due to our paper focused on utilizing VLM-empowered contract theory to assist AIGC task allocation, problem (\ref{Problem_2}) does not take channel, channel power, and computational resource allocation and the associated constraints into consideration. Nonetheless, our proposed method can integrate these into consideration at ease since the VLM-empowered contract theory only adds a constraint to the edge server selection, i.e., narrow the feasible offload set for AIGC task offloading.

\section{Experiment} \label{sec:6}
In this section, we first present the system configuration and experiment benchmarks for our experiment in Section \ref{sec:6.1}. Second, in Section \ref{sec:6.2}, we conduct comprehensive experiments to assess the four evaluation metrics, i.e., the average utility of teleoperators and edge servers and the average response time and completion rate of AIGC tasks, versus the load of the offloading environment. Finally, in Section \ref{sec:6.3}, we conduct experiments to investigate the impact of AIGC task difficulty assessment methods on these four evaluation metrics.

\subsection{Experiment configurations} \label{sec:6.1}

\begin{figure}[!t]
    \begin{center}
        \includegraphics[width=\linewidth]{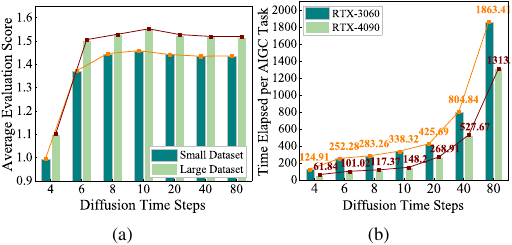}
        \caption{Simulation experiments to identify the computational resource required per AIGC task, where (a) is the evaluation score of processed AIGC task versus the denoising time steps on the AIGC model trained with small and large datasets, (b) is the time elapsed of each AIGC task versus the denoising time steps on RTX-3060 and RTX-4090.}
        \label{fig7}
    \end{center}
\end{figure}

\subsubsection{Experiment Parameters} \label{sec:6.1.1}
We set the number of fog nodes to \{20, 25, 30, 35, 40\}. To estimate the real computational resource required by a low light enhancement AIGC task, we conducted a comprehensive experiment depicted in Fig. \ref{fig7}. Observing Fig. 7a, the average evaluation score of processed AIGC tasks is stabilized after setting the diffusion time steps as 8. Next, Fig. 7b showcased that the time elapsed per AIGC task on RTX-3060 and RTX-4090 are 283.26 ms and 117.37 ms, respectively. Therefore, we assume the computation time per AIGC task on edge servers abides by the uniform distribution of [117.37, 283.26] \footnote{Upon the consideration of the computing power of the edge server is in the range of RTX-3060 and RTX-4090.}.  Similar to \cite{azizi2022deadline}, the transmission delay of the wired connection link between smart gateways and edge servers is set to [1, 3] ms, and the network bandwidth is set to [5, 100] Mbps by referring to \cite{cui2021ol}. We set the number of AIGC tasks to \{100, 150, 200, 250, 300\}. For AIGC tasks, we collected 526 paired low-light and normal-light images via a Unity\footnote{https://unity.com/cn}-based Teleoperation project, in which 80\% training images are used for diffusion-based AIGC model training and 20\% validation images constituted AIGC tasks pool. The AIGC task will be randomly sampled from the AIGC tasks pool. Regarding the AIGC model, we referred to the diffusion-based AIGC model designed in \cite{jiang2023low}. In addition, we set the required response time of AIGC tasks, i.e., $\mathbb{D}$, as 1s, which is in line with the setting presented in \cite{cao2023delay}.

Unless otherwise specified, the VLM generative agents utilized in this paper are based on ChatGPT, and the contract theory-related parameters for simulation are summarized in Table \ref{table1}.

\begin{table}[!t] 
\centering
\caption{Table of Contract Theory-Related Parameters in Simulation}
\label{table1}
\begin{tabular}{|ll|}
\hline
\multicolumn{1}{|c|}{Contract Theory Parameters} & \multicolumn{1}{c|}{Value} \\ \hline
\multicolumn{1}{|c|}{$I_{r1}$}                           & 1.3       \\ \hline
\multicolumn{1}{|c|}{$I_{r2}$}    & 1.4       \\ \hline
\multicolumn{1}{|c|}{$\theta_L$}                       & 1           \\ \hline
\multicolumn{1}{|c|}{$\theta_H$}                     & $\sqrt{2}$       \\ \hline
\multicolumn{1}{|c|}{$\eta_1$}                     & 5          \\ \hline
\multicolumn{1}{|c|}{$\eta_2$}                      & 250              \\ \hline
\multicolumn{1}{|c|}{$\eta_3$}                      & 1               \\ \hline
\multicolumn{1}{|c|}{$\beta_L$}                      & 0.4             \\ \hline
\multicolumn{1}{|c|}{$\beta_H$}                      & 0.6             \\ \hline
\multicolumn{1}{|c|}{$\Delta C$}                           & 10               \\ \hline
\end{tabular}
\vspace{-5mm}
\end{table}

\subsubsection{Benchmarks} \label{sec:6.1.2}
To validate the usefulness of our proposed VLM-empowered contract theory in assisting AIGC task allocation, we compare it with the following benchmark algorithms.
\begin{enumerate}
    \item [(i)] No\_Contract: This comparison benchmark will not adopt contract theory for differential pricing strategy. Therefore, the difficulty level assessment via VLM agents is also not equipped with this benchmark. This benchmark is utilized to validate whether research problem Q1 has been addressed.

    \item [(ii)] Human\_Contract: This method will utilize contract theory to formulate the contract bundle for differential pricing strategy. However, the difficulty level assessment of the AIGC task is conducted by humans \footnote{The assessment result is acquired via voting by five persons, in which each person will rate the difficulty level of the AIGC task by their preference.}. This benchmark is utilized to demonstrate whether research problem Q2 has been tackled.
    
    \item [(iii)] Oracle\_Contract: The contract theory is also utilized in this method, but the difficulty level assessment of the AIGC task is obtained via Proposition \ref{proposition_1}. The role of this benchmark is analogous to Human\_Contract.
\end{enumerate}
To ensure clarity and alignment with the comparison benchmarks, we designate our proposed method to VLM\_Contract.

All the simulations in this paper are based on Python, and the experiments are run on a laptop with a 6-core AMD Ryzen 5, 32 GB of RAM, RTX 4060 GPU, and a Windows 11 operating system. Each experiment was repeated 30 times to enhance the reliability of the result, and the average value was taken as the final result.

\subsection{Impact of Offloading Environment Load} \label{sec:6.2}
\begin{figure}[!t]
\centering
\includegraphics[width=\linewidth]{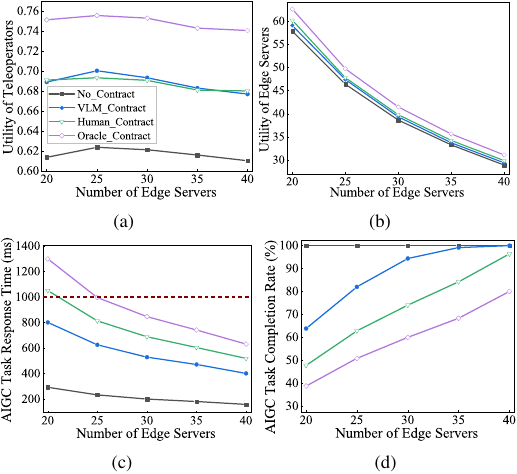}
\caption{Simulation experimental results for fixed AIGC task of 200 and varying the number of edge servers, where (a) is the average utility of teleoperators, (b) is the average utility of edge servers, (c) is the average response time of AIGC task, and (d) is the average completion rate of AIGC task.}
\label{fig8}
\end{figure}

The offloading results will vary as the workload of computing nodes fluctuates. In this section, we will vary the number of edge servers and AIGC tasks, then compare and evaluate the experimental results in four evaluation metrics: the average utility of teleoperators and edge servers and the average response time and completion rate of AIGC tasks.

In the first experiment, we fixed the number of AIGC tasks at 200 and varied the number of edge servers to simulate different loads of offloading environment. As shown in Fig. \ref{fig8}, the average utility of teleoperators is stabilized, the average utility of edge servers and the average response time of AIGC task are decreasing along with the increase of the number of edge servers, and the average completion rate of edge servers is increasing with the number of edge servers increase. The basic rationale is that the computational resource will be ample and the utility of each server will diminish when the number of edge servers rises. Compared with the other three benchmarks, our proposed method, VLM\_Contract, could achieve near-identical performance as Human\_Contract and augment around $10.88 \sim 12.28\% $ and $1.4 \sim 2.17\% $ compared with No\_Contract in terms of the average utility of teleoperators and edge servers, respectively. Notably, as depicted in Figs. 8c and 8d, our proposed method showcased certain deficiencies in terms of the average response time and completion rate of AIGC tasks. However, our proposed method is better than two contract theory-empowered benchmarks, Human\_Contract and Oracle\_Contract. Furthermore, VLM\_Contract could always satisfy the required response time of AIGC tasks and complete all AIGC tasks on time when the computational resource of the offloading environment is ample.

\begin{figure}[!t]
\centering
\includegraphics[width=\linewidth]{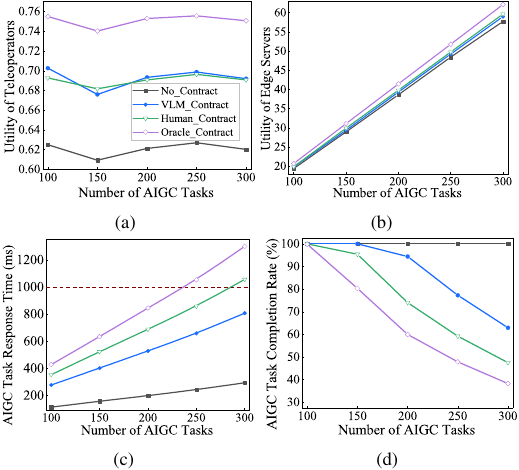}
\caption{Simulation experimental results for a fixed edge server of 30 and varying the number of AIGC tasks, where (a) is the average utility of teleoperators, (b) is the average utility of edge servers, (c) is the average response time of AIGC task, and (d) is the average completion rate of AIGC task.}
\label{fig9}
\end{figure}

In the second experiment, we set the number of edge servers to 30 and varied the number of AIGC tasks. Observing Fig. \ref{fig9}, we could obtain a similar trend as Fig. \ref{fig8} in terms of the four evaluation metrics when the number of AIGC tasks declines. The difference is the average utility of teleoperators and edge servers augmented about $10.94 \sim 12.43\% $ and $1.44 \sim 2.15\% $ when compared with the method of No\_Contract.

\subsection{Impact of Difficulty Assessment Methods} \label{sec:6.3}
In this section, we attempted to investigate the relationship between the methods for difficulty assessment of AIGC tasks and the four evaluation metrics while other parameters are fixed.

By analyzing Fig. 10a, we could conclude that our proposed method can not only assess the AIGC tasks of low difficulty level accurately but also be prone to assign AIGC tasks with uncertainty as high difficulty level, thereby causing teleoperators to possess higher utility in low difficulty type AIGC tasks. As showcased in Fig. 10b, the proposed VLM\_Contract is analogous to No\_Contract in that AIGC tasks with low difficulty would contribute lower utility for edge servers, diverging from the results of Human\_Contract and Oracle\_Contract. The reason for this is two-fold. First, the incentive rationality constraint of the AIGC task in low difficulty levels tends to be breached, causing payment not to be done by teleoperators. Second, more AIGC tasks will be assigned as high difficulty levels, thereby causing edge servers to acquire less utility from low difficulty level AIGC tasks. Observing Figs. 10c and 10d, we could conclude that the load balance of edge servers is worse than the traditional method, i.e., No\_Contract, with the utilization of contract theory. The basic rationale is that the difficulty assessment results of AIGC tasks might not align with the computational resource configuration of the offloading environment.

\begin{figure}[!t]
\centering
\includegraphics[width=\linewidth]{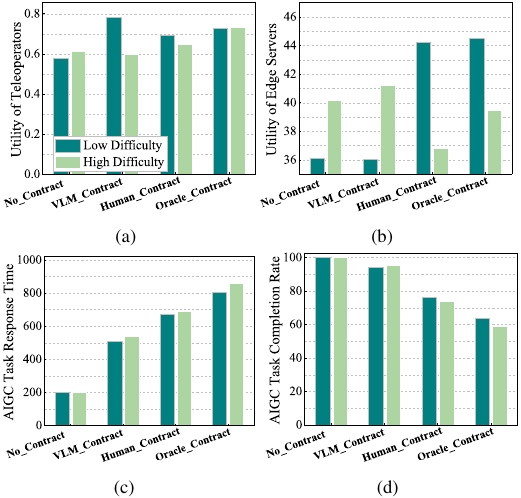}
\caption{Simulation experimental results for assessing the impact of AIGC task difficulty level on four metrics, where (a) is the average utility of teleoperators, (b) is the average utility of edge servers, (c) is the average response time of AIGC task, and (d) is the average completion rate of AIGC task.}
\label{fig10}
\end{figure}

\section{Conclusion} \label{sec:7}
We have introduced a framework of AIGC task allocation assisted by VLM-empowered contract theory in teleoperation. The objectives of our framework are to maximize the average utility of edge servers and guarantee the utility of teleoperators under information asymmetry, i.e., the difficulty of AIGC tasks is hidden information for edge servers. Our devised framework consists of two stages: difficulty assessment of AIGC tasks via VLM-empowered agent and contract theory-assisted AIGC task allocation. In the first stage, we derived the optimal prompt for AIGC task difficulty evaluation as per the capability of chain-of-thought of generative agents, so as to achieve an automatic and accurate AIGC task difficulty assessment. In the second stage, we integrated the difficulty assessment results into the contract bundle formulation and screened the feasible offloading edge server as per the contract selection results of each AIGC task. Experimental results demonstrated that our proposed approach could boost the average utility of teleoperators and edge servers $10.88 \sim 12.43\% $ and $1.4 \sim 2.17\% $, respectively. More specifically, for low-difficulty AIGC tasks, the average utility of teleoperators can augment about $34.84\%$; for high-difficulty AIGC tasks, the average utility of edge servers can improve about $2.77\%$.

\bibliographystyle{IEEEtran}
\bibliography{zhan}

\begin{thebibliography}{10}
\providecommand{\url}[1]{#1}
\csname url@samestyle\endcsname
\providecommand{\newblock}{\relax}
\providecommand{\bibinfo}[2]{#2}
\providecommand{\BIBentrySTDinterwordspacing}{\spaceskip=0pt\relax}
\providecommand{\BIBentryALTinterwordstretchfactor}{4}
\providecommand{\BIBentryALTinterwordspacing}{\spaceskip=\fontdimen2\font plus
\BIBentryALTinterwordstretchfactor\fontdimen3\font minus
  \fontdimen4\font\relax}
\providecommand{\BIBforeignlanguage}[2]{{%
\expandafter\ifx\csname l@#1\endcsname\relax
\typeout{** WARNING: IEEEtran.bst: No hyphenation pattern has been}%
\typeout{** loaded for the language `#1'. Using the pattern for}%
\typeout{** the default language instead.}%
\else
\language=\csname l@#1\endcsname
\fi
#2}}
\providecommand{\BIBdecl}{\relax}
\BIBdecl

\bibitem{zhan2023}
Z.~Zhan, Y.~Dong, D.~M. Doe, Y.~Hu, S.~Li, S.~Cao, W.~Li, and Z.~Han,
  ``Mitigate gender bias in construction: Fusion of deep reinforcement
  learning-based contract theory and blockchain,'' in \emph{IEEE International
  Conference on Blockchain (Blockchain), Danzhou, China}, Dec. 2023.

\bibitem{wang2023improved}
Z.~Wang, Z.~Cai, and Y.~Wu, ``An improved yolox approach for low-light and
  small object detection: Ppe on tunnel construction sites,'' \emph{J. Comput.
  Des. Eng.}, vol.~10, no.~3, pp. 1158--1175, Jun. 2023.

\bibitem{chen2024}
X.~Chen and Y.~Yu, ``An unsupervised low-light image enhancement method for
  improving v-slam localization in uneven low-light construction sites,''
  \emph{Autom. Constr.}, vol. 162, p. 105404, Jun. 2024.

\bibitem{fu2022}
Y.~Fu, Y.~Hong, L.~Chen, and S.~You, ``Le-gan: Unsupervised low-light image
  enhancement network using attention module and identity invariant loss,''
  \emph{Knowl. Based Syst.}, vol. 240, p. 108010, Mar. 2022.

\bibitem{zheng2023learning}
N.~Zheng, J.~Huang, M.~Zhou, Z.~Yang, Q.~Zhu, and F.~Zhao, ``Learning semantic
  degradation-aware guidance for recognition-driven unsupervised low-light
  image enhancement,'' in \emph{Proceedings of the AAAI Conference on
  Artificial Intelligence, Washington DC}, Feb. 2023.

\bibitem{jiang2023low}
H.~Jiang, A.~Luo, H.~Fan, S.~Han, and S.~Liu, ``Low-light image enhancement
  with wavelet-based diffusion models,'' \emph{{ACM} Trans. Graph.}, vol.~42,
  no.~6, pp. 1--14, Dec. 2023.

\bibitem{dhariwal2021diffusion}
P.~Dhariwal and A.~Nichol, ``Diffusion models beat gans on image synthesis,''
  in \emph{Advances in neural information processing systems, virtual-only},
  Dec. 2021.

\bibitem{du2024enabling}
H.~Du, Z.~Li, D.~Niyato, J.~Kang, Z.~Xiong, X.~S. Shen, and D.~I. Kim,
  ``Enabling ai-generated content services in wireless edge networks,''
  \emph{{IEEE} Wirel. Commun.}, vol.~31, no.~3, pp. 226--234, Feb. 2024.

\bibitem{wang2023incentive}
Z.~Wang, Q.~Hu, R.~Li, M.~Xu, and Z.~Xiong, ``Incentive mechanism design for
  joint resource allocation in blockchain-based federated learning,''
  \emph{{IEEE} Trans. Parallel Distributed Syst.}, vol.~34, no.~5, pp.
  1536--1547, May. 2023.

\bibitem{huang2021efficient}
X.~Huang, R.~Yu, D.~Ye, L.~Shu, and S.~Xie, ``Efficient workload allocation and
  user-centric utility maximization for task scheduling in collaborative
  vehicular edge computing,'' \emph{{IEEE} Trans. Veh. Technol.}, vol.~70,
  no.~4, pp. 3773--3787, Apr. 2021.

\bibitem{li2022joint}
Y.~Li, B.~Yang, H.~Wu, Q.~Han, C.~Chen, and X.~Guan, ``Joint offloading
  decision and resource allocation for vehicular fog-edge computing networks: A
  contract-stackelberg approach,'' \emph{{IEEE} Internet Things J.}, vol.~9,
  no.~17, pp. 15\,969--15\,982, Sep. 2022.

\bibitem{zhang2018unreasonable}
R.~Zhang, P.~Isola, A.~A. Efros, E.~Shechtman, and O.~Wang, ``The unreasonable
  effectiveness of deep features as a perceptual metric,'' in \emph{Proceedings
  of the IEEE conference on computer vision and pattern recognition, Salt Lake
  City, UT}, Jun. 2018.

\bibitem{wang2004image}
Z.~Wang, A.~C. Bovik, H.~R. Sheikh, and E.~P. Simoncelli, ``Image quality
  assessment: From error visibility to structural similarity,'' \emph{{IEEE}
  Trans. Image Process.}, vol.~13, no.~4, pp. 600--612, Apr. 2004.

\bibitem{du2023user}
H.~Du, R.~Zhang, D.~Niyato, J.~Kang, Z.~Xiong, S.~Cui, X.~Shen, and D.~I. Kim,
  ``User-centric interactive ai for distributed diffusion model-based
  ai-generated content,'' \emph{arXiv preprint arXiv:2311.11094}, vol.
  abs/2311.11094, Nov. 2023.

\bibitem{wang2024guiding}
J.~Wang, H.~Du, D.~Niyato, Z.~Xiong, J.~Kang, S.~Mao, and X.~S. Shen, ``Guiding
  ai-generated digital content with wireless perception,'' \emph{{IEEE} Wirel.
  Commun.}, pp. 1--8, Apr. 2024, early Access.

\bibitem{fan2023learning}
J.~Fan, M.~Xu, Z.~Liu, H.~Ye, C.~Gu, D.~Niyato, and K.-Y. Lam, ``A
  learning-based incentive mechanism for mobile aigc service in decentralized
  internet of vehicles,'' in \emph{IEEE 98th Vehicular Technology Conference
  (VTC2023-Fall), HongKong, China}, Oct. 2023.

\bibitem{du2024diffusion}
H.~Du, Z.~Li, D.~Niyato, J.~Kang, Z.~Xiong, H.~Huang, and S.~Mao,
  ``Diffusion-based reinforcement learning for edge-enabled ai-generated
  content services,'' \emph{IEEE Trans. Mob. Comput.}, pp. 1--16, Jan. 2024,
  early Access.

\bibitem{du2023exploring}
H.~Du, R.~Zhang, D.~Niyato, J.~Kang, Z.~Xiong, D.~I. Kim, X.~S. Shen, and H.~V.
  Poor, ``Exploring collaborative distributed diffusion-based ai-generated
  content (aigc) in wireless networks,'' \emph{{IEEE} Netw.}, vol.~38, no.~3,
  pp. 178--186, May. 2024.

\bibitem{liu2024blockchain}
Y.~Liu, H.~Du, D.~Niyato, J.~Kang, Z.~Xiong, C.~Miao, X.~S. Shen, and
  A.~Jamalipour, ``Blockchain-empowered lifecycle management for ai-generated
  content products in edge networks,'' \emph{{IEEE} Wirel. Commun.}, vol.~31,
  no.~3, pp. 286--294, Jun. 2024.

\bibitem{lin2023blockchain}
Y.~Lin, Z.~Gao, H.~Du, and D.~Niyato, ``Blockchain-aided ai-generated content
  services: Stackelberg game-based content caching approach,'' in \emph{IEEE
  International Conference on Web Services (ICWS), Chicago, IL}, Jul. 2023.

\bibitem{wang2023security}
T.~Wang, Y.~Zhang, S.~Qi, R.~Zhao, Z.~Xia, and J.~Weng, ``Security and privacy
  on generative data in aigc: A survey,'' \emph{arXiv preprint
  arXiv:2309.09435}, vol. abs/2309.09435, Dec. 2023.

\bibitem{zhang2017contract}
Y.~Zhang and Z.~Han, \emph{Contract Theory for Wireless Networks}.\hskip 1em
  plus 0.5em minus 0.4em\relax Springer, Feb. 2017.

\bibitem{dang2023contract}
T.~N. Dang, A.~Manzoor, Y.~K. Tun, S.~A. Kazmi, Z.~Han, and C.~S. Hong, ``A
  contract theory-based incentive mechanism for uav-enabled vr-based services
  in 5g and beyond,'' \emph{{IEEE} Internet Things J.}, vol.~10, no.~18, pp.
  16\,465--16\,479, Sep. 2023.

\bibitem{zheng2023contract}
Y.~Zheng, L.~Zou, W.~Zhang, J.~Yang, L.~Yang, and Z.~Lin, ``Contract-based
  cooperative computation and communication resources sharing in mobile edge
  computing,'' \emph{J. Grid Comput.}, vol.~21, no.~1, pp. 1--19, Feb. 2023.

\bibitem{diamanti2022incentive}
M.~Diamanti, P.~Charatsaris, E.~E. Tsiropoulou, and S.~Papavassiliou,
  ``Incentive mechanism and resource allocation for edge-fog networks driven by
  multi-dimensional contract and game theories,'' \emph{{IEEE} Open J. Commun.
  Soc.}, vol.~3, no.~1, pp. 435--452, Feb. 2022.

\bibitem{shen2023dynamic}
R.~Shen, M.~Gao, W.~Li, and Y.~Li, ``Dynamic task offloading in distributed vec
  networks: An exploration and exploitation assisted contract-theoretic
  approach,'' \emph{{IEEE} Trans. Veh. Technol.}, vol.~73, no.~4, pp.
  5717--5729, Apr. 2024.

\bibitem{yang2020resource}
C.~Yang, W.~Lou, Y.~Liu, and S.~Xie, ``Resource allocation for edge
  computing-based vehicle platoon on freeway: A contract-optimization
  approach,'' \emph{{IEEE} Trans. Veh. Technol.}, vol.~69, no.~12, pp.
  15\,988--16\,000, Dec. 2020.

\bibitem{wen2023task}
J.~Wen, J.~Kang, Z.~Xiong, Y.~Zhang, H.~Du, Y.~Jiao, and D.~Niyato, ``Task
  freshness-aware incentive mechanism for vehicle twin migration in vehicular
  metaverses,'' in \emph{IEEE International Conference on Metaverse Computing,
  Networking and Applications (MetaCom), Kyoto, Japan}, Jun. 2023.

\bibitem{kazmi2021novel}
S.~A. Kazmi, T.~N. Dang, I.~Yaqoob, A.~Manzoor, R.~Hussain, A.~Khan, C.~S.
  Hong, and K.~Salah, ``A novel contract theory-based incentive mechanism for
  cooperative task-offloading in electrical vehicular networks,'' \emph{{IEEE}
  Trans. Intell. Transp. Syst.}, vol.~23, no.~7, pp. 8380--8395, Jul. 2022.

\bibitem{cao2023delay}
S.~Cao, Z.~Zhan, C.~Dai, S.~Chen, W.~Zhang, and Z.~Han, ``Delay-aware and
  energy-efficient iot task scheduling algorithm with double blockchain enabled
  in cloud-fog collaborative networks,'' \emph{{IEEE} Internet Things J.},
  vol.~11, no.~2, pp. 3003--3016, Jan. 2024.

\bibitem{ho2020denoising}
J.~Ho, A.~Jain, and P.~Abbeel, ``Denoising diffusion probabilistic models,''
  \emph{Advances in neural information processing systems, virtual-only}, Dec.
  2020.

\bibitem{song2021denoising}
J.~Song, C.~Meng, and S.~Ermon, ``Denoising diffusion implicit models,'' in
  \emph{9th International Conference on Learning Representations (ICLR),
  virtual-only}, May. 2021.

\bibitem{azizi2022deadline}
S.~Azizi, M.~Shojafar, J.~Abawajy, and R.~Buyya, ``Deadline-aware and
  energy-efficient iot task scheduling in fog computing systems: A semi-greedy
  approach,'' \emph{J. Netw. Comput. Appl.}, vol. 201, p. 103333, May. 2022.

\bibitem{wei2022chain}
J.~Wei, X.~Wang, D.~Schuurmans, M.~Bosma, F.~Xia, E.~Chi, Q.~V. Le, D.~Zhou
  \emph{et~al.}, ``Chain-of-thought prompting elicits reasoning in large
  language models,'' \emph{Advances in neural information processing systems,
  New Orleans, LA}, Nov. 2022.

\bibitem{li2021contract}
J.~Li, T.~Liu, D.~Niyato, P.~Wang, J.~Li, and Z.~Han, ``Contract-theoretic
  pricing for security deposits in sharded blockchain with internet of things
  (iot),'' \emph{{IEEE} Internet Things J.}, vol.~8, no.~12, pp.
  10\,052--10\,070, Jun. 2021.

\bibitem{cui2021ol}
G.~Cui, Q.~He, X.~Xia, F.~Chen, F.~Dong, H.~Jin, and Y.~Yang, ``Ol-eua: Online
  user allocation for noma-based mobile edge computing,'' \emph{{IEEE} Trans.
  Mob. Comput.}, vol.~22, no.~4, pp. 2295--2306, Apr. 2021.

\end{thebibliography}
\end{document}